\documentclass{sn-jnl}
\usepackage{xcolor}
\usepackage{graphicx}
\usepackage{array}
\usepackage{multirow}
\usepackage{amsmath, amssymb, amsfonts}
\usepackage{amsthm}
\usepackage{mathrsfs}
\usepackage[title]{appendix}
\usepackage{textcomp}
\usepackage{manyfoot}
\usepackage{booktabs}
\usepackage{algorithm, algpseudocode, algorithmicx} 
\usepackage{listings}
\usepackage{soul}
\usepackage[caption=false, font=normalsize, labelfont=sf, textfont=sf]{subfig}
\usepackage{balance}
\usepackage{url}
\usepackage{verbatim}
\usepackage{stfloats}
\usepackage{hyperref}
\usepackage[authoryear, round]{natbib} 

\begin{document}

\title[Article Title]{Refined Causal Graph Structure Learning via Curvature for Brain Disease Classification}


\author[1]{\fnm{Falih Gozi} \sur{Febrinanto}}\email{f.febrinanto@federation.edu.au}
\author[2]{\fnm{Adonia} \sur{Simango}}\email{asimango@gmail.com}
\author[3]{\fnm{Chengpei} \sur{Xu}}\email{chengpei.xu@unsw.edu.au}
\author*[4]{\fnm{Jingjing} \sur{Zhou}}\email{zhoujingjing@zjgsu.edu.cn}
\author[1]{\fnm{Jiangang} \sur{Ma}}\email{j.ma@federation.edu.au}
\author[5]{\fnm{Sonika} \sur{Tyagi}}\email{sonika.tyagi@rmit.edu.au}
\author*[5]{\fnm{Feng} \sur{Xia}}\email{f.xia@ieee.org}

\affil[1]{\orgdiv{Institute of Innovation, Science, and Sustainability}, \orgname{Federation University Australia}, \orgaddress{\city{Ballarat}, \postcode{3353}, \state{VIC}, \country{Australia}}}

\affil[2]{\orgname{Stainaz Research}, \orgaddress{\city{Johannesburg}, \postcode{2192}, \state{Gauteng}, \country{South Africa}}}

\affil[3]{\orgdiv{MIoT \& IPIN Lab}, \orgname{University of New South Wales}, \orgaddress{\city{Sydney}, \postcode{2052}, \state{NSW}, \country{Australia}}}

\affil[4]{\orgdiv{School of Information and Electronic Engineering}, \orgname{Zhejiang Gongshang University}, \orgaddress{\city{Hangzhou}, \postcode{310018}, \country{China}}}

\affil[5]{\orgdiv{School of Computing Technologies}, \orgname{RMIT University}, \orgaddress{\city{Melbourne}, \postcode{3000}, \state{VIC}, \country{Australia}}}


\abstract{Graph neural networks (GNNs) have been developed to model the relationship between regions of interest (ROIs) in brains and have shown significant improvement in detecting brain diseases. However, most of these frameworks do not consider the intrinsic relationship of causality factor between brain ROIs, which is arguably more essential to observe cause and effect interaction between signals rather than typical correlation values. We propose a novel framework called CGB (Causal Graphs for Brains) for brain disease classification/detection, which models refined brain networks based on the causal discovery method, transfer entropy, and geometric curvature strategy. CGB unveils causal relationships between ROIs that bring vital information to enhance brain disease classification performance. Furthermore, CGB also performs a graph rewiring through a geometric curvature strategy to refine the generated causal graph to become more expressive and reduce potential information bottlenecks when GNNs model it. Our extensive experiments show that CGB outperforms state-of-the-art methods in classification tasks on brain disease datasets, as measured by average F1 scores.}



\maketitle

\section{Introduction} 
The field of neuroscience has been revolutionized by the advent of brain imaging technologies, particularly functional magnetic resonance imaging in the resting state (rest fMRI)~\citep{khalilullah2023multimodal_fmri_intro, vasilkovska2023resting_fmri_intro, liu2024motif_neuro}. This powerful tool allows the measurement of blood-oxygen-level-dependent (BOLD) signals in predefined Regions of Interest (ROIs) within the brain, offering an unprecedented avenue for revealing information about potential diseases such as autism spectrum disorder (ASD) and schizophrenia~\citep{philiastides2021inferring_bold_intro, kocak2021artificial_bold_intro}. Various brain atlases, including Harvard-Oxford~\citep{makris2006decreased_harvard_Parcellation} and Craddock 200~\citep{craddock2012whole_cradock_Parcellation} parcellations, have been used to define these ROIs. Furthermore, ROIs can be interestingly modelled as graph data, where the ROIs themselves represent nodes, and the connections between ROIs represent edges of graphs~\citep{cui2022interpretable_brain_as_graph}. This graph-based data structure, inheriting the graph theory technique, has been instrumental in revealing meaningful relationships between ROIs in brain networks to diagnose brain diseases more effectively~\citep{alsubaie2024alzheimer_newsurvey, ren2024brain_newsurvey}.

With the current popularity of deep learning, recent frameworks have developed graph neural networks (GNNs)~\citep{xia2021graph, febrinanto2023graph} to extend the merits of modelling graph-structured data for detecting brain diseases with brain networks based on fMRI signals as input~\citep{kan2022brain_brain_transformer, li2021braingnn,kan2022fbnetgen,cui2022braingb,elgazzar2022benchmarking, febrinanto2023balanced}. These techniques perform more accurately than typical machine learning or deep learning techniques. However, there is still a high consensus on how to construct or define an appropriate graph structure in brain networks in terms of two processes: 1) how do we generate the graphs? and 2) How do we refine the graphs? 

\bmhead{Current Limitations} To generate graph structures from scratch, some methods utilize learnable techniques to create optimal structures over different variations of sample data through an end-to-end learning process~\citep{kan2022fbnetgen, kazi2022differentiable_dgm,kan2022brain_brain_transformer}. Even though this technique achieves substantial performance in detecting brain diseases, it fails to incorporate biological insight or actual domain knowledge of fMRI signals, making it lack generalization to unseen data samples and prone to overfitting problems driven solely by the available data samples~\citep{chen2023balanced_learnable_lacks}. These problems of lack of generalization and overfitting significantly reduce the performance of brain disease classification. Other methods adopt a correlation matrix to represent the graph structure in brain networks, calculating the similarity between series of BOLD signals across all brain regions~\citep{cui2022braingb,elgazzar2022benchmarking,li2021braingnn}. This technique integrates biological insights or domain-specific knowledge about fMRI signals within each ROI to construct graphs. However, it neglects critical cause-and-effect relationships in fMRI signal connections~\citep{assaad2022survey_causal_relationship}, failing to determine whether one ROI signal influences another. Causality is crucial as it reveals information flow between signal pairs, reflecting more meaningful interactions between brain regions. These interaction patterns of causal networks may differ significantly between healthy individuals and those with disease conditions such as Alzheimer’s disease or Attention-Deficit/Hyperactivity Disorder (ADHD)~\citep{cui2022braingb}. Comparing the brain network patterns can provide valuable insights into improving the performance of brain disease classification.

A graph refinement, also known as graph rewiring, is crucial for further enhancing the expressiveness of the graph structure, reducing noisy elements, and mitigating potential bottleneck issues while modeling it with Graph Neural Networks (GNNs)~\citep{akansha2023over_rewiring, li2024addressing_oversquashing}. This adoption is still in its infancy in GNNs for brain disease classification. The most straightforward refinement method is to conduct data-driven analysis on thresholded connectivity in the brain networks~\citep{cui2022braingb,li2021braingnn}. This thresholding technique aims to reduce the complexity of the graph structure and improve computation time by removing less meaningful relationships. However, it can lead to the inclusion of spurious connections, confounding the analysis of the relationship between ROIs. \cite{elgazzar2022benchmarking} employ a graph diffusion technique~\citep{gasteiger2019diffusion_dgc} to address noisy graphs. This technique deletes and adds edges to produce an optimal graph representation. However, this refinement strategy tends to alter the original characteristics of the graphs and yield a relatively different view of the graph structure that shifts the knowledge and information from the original graph structure~\citep{topping2021understanding_curvature}.

\bmhead{Our Work} To address the current challenge of generating and refining graph structure, we propose CGB (Causal Graphs for Brains) for brain disease classification. Our objective is to pre-construct the graph structure based on causal relationships between fMRI signals of the brain's ROIs, thereby enabling causality factors to explain whether a signal in an ROI can cause another signal. We introduce a strategy that leverages the causal discovery of multiple signals using transfer entropy~\citep{schreiber2000measuring_transferentropy}. Transfer entropy is a comprehensive information theory-based approach that uncovers causality among variables in multiple series signals. It measures the amount of information transferred from one variable to another over time, enabling graph structure generation with edges representing the causal relationships between variables, in this case, ROIs~\citep{jiao2020quality_causality}. This helps add significant insight into the information flow between signals, enhancing brain disease classification performance.

For addressing the graph refinement challenge, CGB adopts graph rewiring via the \textit{geometry curvature technique}~\citep{topping2021understanding_curvature}. CGB's graph refinement strategy aims to make the representation of the graph more expressive, reduce the noisy elements of graph structures, and address the information bottleneck problem in GNNs. Unlike previous methods of graph refinement in brain networks, our model can maintain the characteristics of the original graph without drastically changing the distribution of the structures~\citep{topping2021understanding_curvature}. Thus, \textit{it preserves the original causal information} from the original graph and creates the optimal structure of brain networks.

\bmhead{Contributions} In summary, this paper presents four main contributions:
\begin{itemize}
\item We propose a novel framework for brain disease classification called CGB. This framework effectively captures intrinsic causal relationships between ROIs in the brain, providing more informative knowledge and offering GNN modelling to enhance brain disease classification performance.

\item We present a refined causal graph generation method. It aims to develop a graph adjacency matrix with a causality factor based on transfer entropy and optimize the obtained causal graph via \textit{the geometry curvature technique}.

\item We devise a module called Causality-Informed Stochastic Discrete Ricci Flow (CSDRF), which improves the geometric curvature strategy to preserve the causal relationships between nodes from drastic changes during the graph refinement process.

\item We conduct extensive experiments on several brain disease datasets. Our experimental results show that CGB outperforms baseline models in detecting diverse brain diseases.
\end{itemize}

\section{Related Work}
\subsection{GNNs for Brain Disease Detection}
Recent works have proposed to harness the potential of GNNs for brain network analysis. For example, \cite{elgazzar2022benchmarking} uses correlation calculation to create a graph structure and GNNs to model the graph structure. \cite{li2021braingnn} in their framework called BrainGNN develops ROI-aware GNNs to utilize a particular pooling strategy to select important nodes. On the other hand, \cite{kan2022fbnetgen} proposed FBNetGen that uses a learnable graph for creating brain networks and investigating the interpretability of the generated networks for downstream tasks. Besides, \cite{kazi2022differentiable_dgm} introduced DGM, which designs a latent-graph learning block to build a probabilistic graph. BrainNETTF by \cite{kan2022brain_brain_transformer} employs graph transformers to model brain networks and uses a clustering readout strategy to perform classification. Bargain by \cite{febrinanto2023balanced} generates multiview graph structures from brains to perform disease detection. The most recent approach, TSEN by \cite{hu2023transformer_TSEN}, proposes a transformer and snowball graph convolution learning for brain functional network classification.

\subsection{Graph Rewiring}
Graph rewiring aims to enhance the expressiveness of the graph structure, reduce noisy elements, and improve GNNs' ability in task-specific learning~\citep{akansha2023over_rewiring, fabbri2022rewiring_rewiring}. The graph rewiring framework differs from GNNs that aim to improve performance by learning the graph structure based on the tasks. Graph rewiring is a process of changing the graph structure to control the information flow and enhance the ability of the network to perform tasks~\citep{arnaiz2022diffwire_rewiringIntro, li2024addressing_oversquashing}. Several approaches for graph rewiring have been proposed, including DIGL introduced by \cite{gasteiger2019diffusion_dgc} that utilizes the diffusion-based preprocessing method. \cite{topping2021understanding_curvature} proposed SDRF (Stochastic Discrete Ricci Flow) that performs geometry curvature rewiring to optimize the structures. The work of \cite{arnaiz2022diffwire_rewiringIntro} called DIFFWIRE implements both diffusion and curvature techniques. GTR by \cite{black2023understanding_GTR} uses a rewiring technique based on effective resistance that explains the commute time between two vertices.

In GNNs for brain disease classification, the most straightforward technique to perform graph network rewiring by conducting data-driven analysis is using thresholding techniques~\citep{cui2022braingb,li2021braingnn} to reduce the complexity of the graph structure in brains. However, this trivial technique can only remove edges that can cause loss of spurious connections. GDC-GCN by \cite{elgazzar2022benchmarking} for brain disease classification uses the graph diffusion based on DIGL~\citep{gasteiger2019diffusion_dgc} technique to address noisy graphs. However, this refinement strategy tends to alter the original characteristics of graphs and can drastically change the graph structure from its original view.

\subsection{Causal Discovery}
Causal discovery is a technique to understand causal factors between variables in signals or time series data. Unlike simple correlation analysis, revealing causal relationships between variables is essential for gaining insights and further optimizing data analysis techniques. Granger causality is a popular method for measuring causal relationships between time series data~\citep{granger1969investigating_gragercausality}. This approach uses an auto-regression method to test the statistical hypothesis of causality, but it mainly captures linear relationships. Transfer entropy is another method considered generalized Granger causality~\citep{schreiber2000measuring_transferentropy}. Unlike Granger causality, transfer entropy is adept at capturing both linear and nonlinear relationships~\citep{jiao2020quality_causality}. 

Causal discovery using transfer entropy can be use to reveal dependencies between signals, representing them as a graph where nodes are signals and edges are dependencies. This type of graph have been used in various tasks such as forecasting~\citep{duan2022multivariate_causalGraph} and anomaly detection in cyber-physical systems~\citep{febrinanto2023entropy_causal_graph}. We extend this by integrating causal discovery and transfer entropy with a graph rewiring strategy based on geometric curvature, refining graph structures to enhance expressiveness and address bottlenecks when modeled with GNNs~\citep{topping2021understanding_curvature}.

\section{Preliminaries}
In this work, each brain's ROI produces a time series signal. A collection of overall ROIs' time series signals can be represented by a matrix $V \in \mathbb{R}^{n \times t}$ where $n$ is the number of nodes (ROIs), and $t$ represents a period of recording time.

\bmhead{Brain's Graph Structure} A brain network can be represented as a graph structure denoted by $G=(A, X)$, where $A \in \mathbb{R}^{n \times n}$ represents the causal graph adjacency matrix and $X \in \mathbb{R}^{n \times n}$ represents the features of nodes. Here, $n$ denotes the number of nodes. The node features $X$ are derived from causality calculations between ROIs $\textbf{v}_i \in \mathbb{R}^{t}$ and $\textbf{v}_j \in \mathbb{R}^{t}$.

\bmhead{Brain Disease Classification/Detection} This work uses GNN techniques for brain disease/disorder detection, which is a classification task. First, we apply GNNs to perform node embedding on the graph $G=(A, X)$, resulting in new node embeddings after several layers denoted as $Z\in \mathbb{R}^{n \times f}$, where $f$ is the size of GNN embeddings. Subsequently, we use a graph pooling mechanism to transform the node embedding $Z$ into a graph-level embedding represented by vector $\textbf{z}^{\textup{pooled}} \in \mathbb{R}^m$, where $m$ is the size of pooling vector. Finally, the graph-level embedding vector $\textbf{z}^{\textup{pooled}}$ is passed to fully connected network layers as feature transform for making graph-level predictions, $y$. In that case, when the value of $y$ is 1, indicating a specific brain disease, or when the value of $0$, it indicates a normal sample.

\begin{figure}[t]
  \centering
  \includegraphics[width=0.9\textwidth]{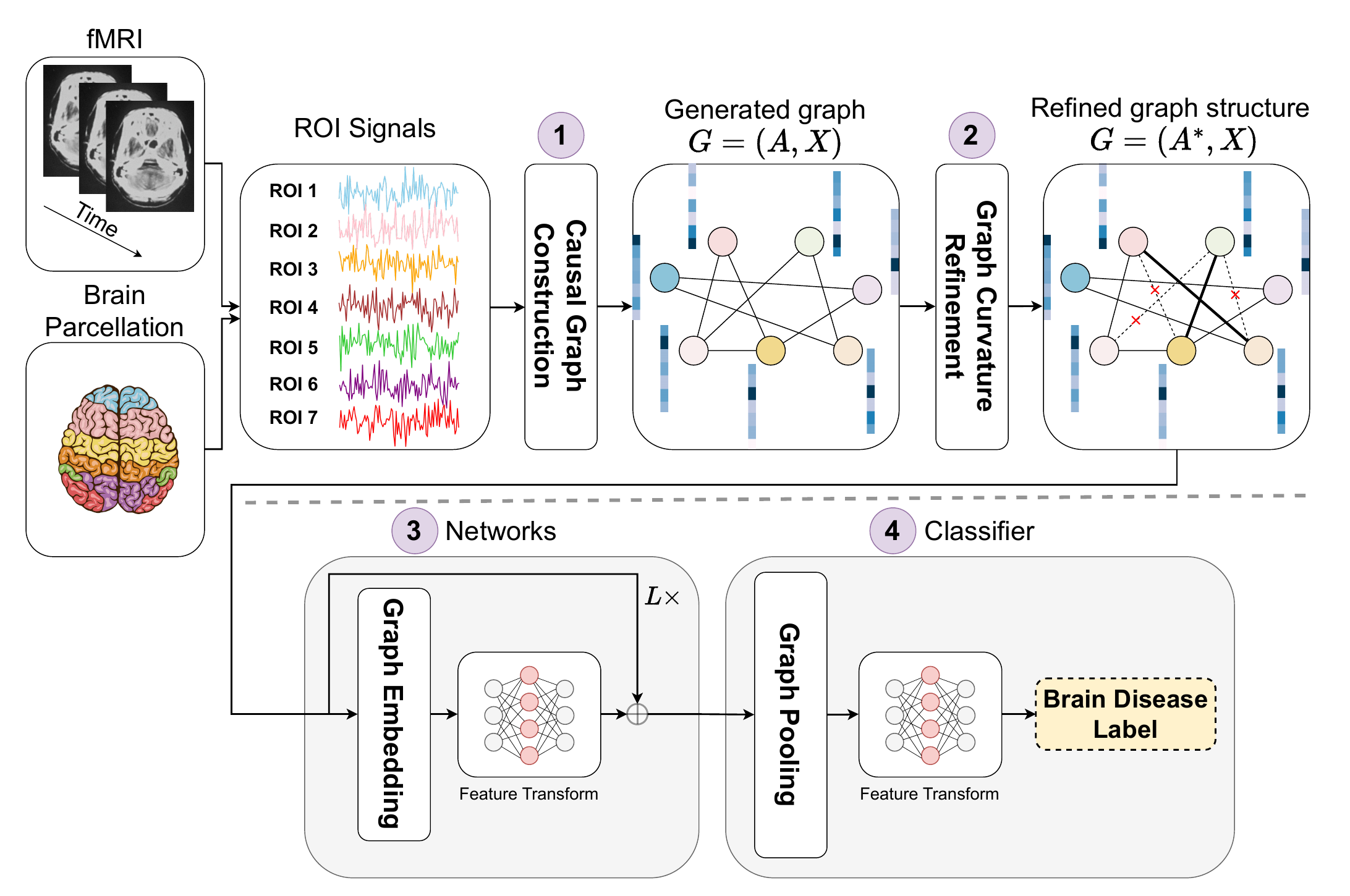}
  \caption{Overview of the proposed CGB framework. The process begins with \textcircled{1} constructing a causal graph from ROI signals in the brain using Transfer Entropy. Next, \textcircled{2} graph curvature refinement enhances the structural expressiveness of the causal graph. Then, \textcircled{3} a graph embedding process leverages GNNs to model the refined graph. Finally, \textcircled{4} a pooling mechanism transforms the graph embedding into a high-level representation, which is then mapped to a specific brain disease label.}
  \label{fig:framework}
\end{figure}

\section{Methodology}

Our framework, CGB for brain disease classification, is divided into three main parts:

\begin{itemize}
\item \textbf{Refined causal graph generation:} to generate graph structure using causal discovery based on transfer entropy and graph refinement strategy based on the geometry curvature technique,
\item \textbf{Brain's spatial modelling:} to learn the spatial representation of brain networks given the refined graph structure with GNNs,
\item \textbf{Pooling and classifier modules:} to summarize every obtained node representation in the graph by combining it into graph-level representation since disease classification is a graph-level task and to map the graph-level representation into binary classification to categorize normal or diseased brains (Fig.~\ref{fig:framework}).
\end{itemize}

\subsection{Refined Causal Graph Generation}
In this section, we first introduce a brain network generation method using transfer entropy (TE). Transfer entropy is a measure within information theory, based on Shannon entropy's conditional probabilities. It measures the causality factor between two time-series signals in the brain regions of interest (ROIs). TE has been demonstrated to have superior results and visual interpretability on causality calculation compared to other causal discovery methods like Granger Causality~\citep{granger1969investigating_gragercausality}. This brain's graph generation module based on TE aims to construct a graph adjacency matrix for brain networks. For example, considering two signals from the brain region, $\textbf{v}_i \in \mathbb{R}^{t}$ and $\textbf{v}_j \in \mathbb{R}^{t}$, we generate directed relationships between these regions when there is causal influence from $\textbf{v}_i$ to $\textbf{v}_j$ or $\textbf{v}_j$ to $\textbf{v}_i$.

\bmhead{TE Calculation}
The TE formula begins by generating a histogram to estimate each time series's probability density function (PDF). The number of bins, denoted as $D$, is determined based on the ranges of all possible values in a given (univariate) time series. Subsequently, the values in each time series, such as $v_i$, are assigned to their corresponding bins based on the associated data points. The Shannon entropy, denoted as $H(I)$, can be calculated as:

\begin{equation}
  \label{eq:te1}
  H(I) = -\sum_{i\in I} p(i)\log_2 p(i),
\end{equation}
In this context, $p(\cdot)$ represents the probability density function for the variable $I$, which takes values in the range $\{1,\dotsc, D\}$. This variable measures the position of the histogram value within the $D$ bins, based on the time series $v_{i}$ values it falls into. The same definition of Shannon entropy can also be implemented for multiple variables called joint entropy. Consider another variable $J\in  \{1,..., D\}$ corresponding to time series $v_j$. The joint Shannon entropy $H(I,J)$ is defined as:

\begin{equation}
  \label{eq:te2}
  H(I,J) = -\sum_{i \in I, j \in J} p(i,j)\log_2 p(i,j),
\end{equation} 

Apart from the joint entropy $H(I, J)$, another important term in information theory is the conditional entropy $H(I|J)$. This term quantifies the amount of information in variable $I$ when variable $J$ is known:

\begin{equation}
  \label{eq:te3}
  H(I|J) = -\sum_{i \in I, j \in J} p(i,j)\log_2 p(i|j),
\end{equation}

TE, represented as $T_{J\rightarrow I}$, measures the information flow from variable $J$ to variable $I$~\citep{schreiber2000measuring_transferentropy}:

\begin{equation}
  \label{eq:te4}
  T_{J\rightarrow I} = H(I_{t+1}|I_{t}) - H(I_{t+1}|I_{t},J_{t}),
\end{equation}

Here, the indices $t+1$ and $t$ represent consecutive time points. The complete equation for TE is as follows:

\begin{equation}
  \label{eq:te5}
  T_{J\rightarrow I} = \sum p(i_{t}, \textbf{i}_{t-1}^{(q)}, \textbf{j}_{t-1}^{(o)})\log_2\frac{p(i_t|\textbf{i}_{t-1}^{(q)},\textbf{j}_{t-1}^{(o)})}{p(i_t|\textbf{i}_{t-1}^{(q)})},
\end{equation}
where $\textbf{i}_{t-1}^{(q)}$ represents all possible histogram encodings based on time series $\textbf{v}_i$, with the latest value at time $t-1$ included. The time series $\textbf{v}_i$ is defined as $\textbf{v}_i = [v_{i;t-1}, v_{i;t-2}, \ldots, v_{i;t-q+1}]$. Similarly, $\textbf{j}_{t-1}^{(o)}$ comprises all possible histogram encodings based on time series $\textbf{v}_j$, with the latest value at time $t-1$ included. Time series $\textbf{v}_j$ is defined as $\textbf{v}_j = [v_{j;t-1}, v_{j;t-2}, \ldots, v_{j;t-o+1}]$. For the window size $q$, the most natural choices are either $o$ or $1$ as suggested in~\cite{schreiber2000measuring_transferentropy}.

\bmhead{Graph Structure Utilization} Several works have focused on utilizing TE to develop graphs~\citep{duan2022multivariate_causalGraph, febrinanto2023entropy_causal_graph}. In this work, we specifically use TE calculations to generate a graph structure in the brain network, resulting in a directed adjacency matrix developed as follows:

\begin{equation}
  \label{eq:adj}
  A_{ij} =
  \begin{cases}
    1 & \text{if $T_{\textbf{v}_j\rightarrow \textbf{v}_i} > c;$} \\
    0           & \text{otherwise},
  \end{cases}
\end{equation}
let $c$ be a control constant to prevent weak relations from being developed. The value of $c$ can be selected as a hyperparameter.

\begin{figure}[ht]
\centering
\includegraphics[width=0.75\textwidth]{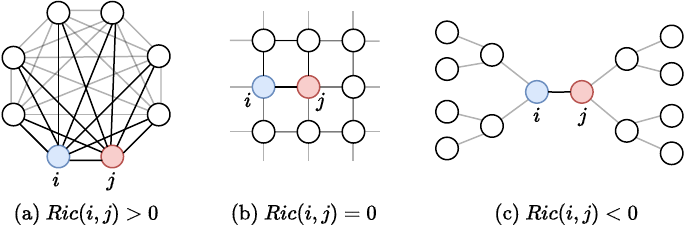}
\caption{Illustration of edges in a graph with different curvature~\citep{topping2021understanding_curvature}.}
\label{fig:curvature}
\end{figure}

\bmhead{Curvature Graph Refinement} After we obtain the initial graph structure of brains, graph refinement or rewiring is required to enhance the expressiveness of the graph structure. 
The TE calculation might not be accurate when noisy signals appear in fMRI data due to scanner drift, physiological noise (e.g., cardiac pulsation), or body motion ~\citep{hutchison2013dynamic_issues}. Moreover, some topology patterns in the graph structure can cause a potential bottleneck problem~\citep{li2024addressing_oversquashing}. The information bottleneck causes GNNs to be unable to capture interactions between distant nodes. Thus, it is important to alter (i.e., rewire) the original graph as a preprocessing to improve the GNN process. In this work, we adopt and improve the balanced Forman curvature of SDRF (Stochastic Discrete Ricci Flow)~\citep{topping2021understanding_curvature} framework, which theoretically justifies local topology that resembles a tree that can cause information bottlenecks in graph structures. It states that resolving a tree topology in graph structure improves the graph expressiveness. We further enhance this graph refinement strategy to ensure that the refined graph preserves the causal relationships between nodes, as discussed later in this section.

The balanced Forman curvature is based on a particular geometry theory, Ricci curvature. As shown in Fig.~\ref{fig:curvature}, given a pair of nodes $i$ and $j$ connected by an edge $e_{ij}$, we can calculate the curvature $Ric(i,j)$ based on surrounding topological connections such as triangles and four cycles based at edge$(i,j)$. A higher curvature implies that information could be propagated more easily from $i$ to $j$ and vice versa (as shown in Fig.~\ref{fig:curvature} (a)), and this should be preserved. In contrast, the edges with low curvature indicate little information flows from $i$ to $j$ (as shown in Fig.~\ref{fig:curvature} (c)) that can cause information bottlenecks. Thus, it should be refined by adding some edges to improve the expressiveness of the graph structure. Formally, the curvature is defined as follows:

\begin{equation}
  \label{eq:curv}
  \begin{aligned}
  \textup{Ric}(i,j) = \frac{2}{d_i} + \frac{2}{d_j} - 2 + 2\frac{|\#_{\triangle}(i,j)|}{\textup{max}\{d_i,d_j\}} + \frac{|\#_{\triangle}(i,j)|}{\textup{min}\{d_i,d_j\}} \\
  + \frac{(\gamma_{\textup{max}})^{-1}}{\textup{max}(d_i,d_j)}(|\#_{\square}(i,j)| + |\#_{\square}(j,i)|),
  \end{aligned}
\end{equation}
where $d_i$ represents degree of a node, $\#_{\triangle}(i,j)$ is the number of triangles on the edge $(i,j)$, $\#_{\square}(i,j)$ is the number of neighbours of node $i$ forming a 4-cycle based on the edge $(i,j)$ without diagonals inside, and $\gamma_{\textup{max}}$ is the maximal number of 4-cycles based at edge $(i,j)$ traversing a common node.

The balanced Forman curvature proposed in SDRF~\citep{topping2021understanding_curvature} only considers the improvement of the curvature factor when introducing new edges. Then, it chooses one of the best edges as a new relation in the graph to optimize the structure. However, this strategy can not be directly implemented in a causal graph since the relationship in the causal graph brings a causal factor that needs to be considered. Only considering curvature improvement can introduce a relationship with \textit{no causal effect}. 

\begin{algorithm}[ht]
  \caption{Causality-Informed Stochastic Discrete Ricci Flow (CSDRF)}
  \label{alg:sdrf}
  \begin{algorithmic}[1]
    \Require 
    \Statex Causal graph adjacency matrix $A$, temperature $\tau > 0$, maximum iterations, and optional bounds $C^+$ and $C^-$.
    
    \Repeat
      \State \textbf{Identify Bottleneck Edge:} Find the edge $(i, j)$ with the lowest Ricci curvature $\textup{Ric}(i, j)$.
      \State \textbf{Compute Improvements:} For each candidate edge $(k, l)$ with $k \in B_1(i)$ and $l \in B_1(j)$, compute 
      \[
      x_{kl} = (\textup{Ric}_{kl}(i, j) - \textup{Ric}(i, j)) \cdot T_{\textbf{v}_k \rightarrow \textbf{v}_l}.
      \]
      \State \textbf{Sample New Edge:} Sample edge $(k, l)$ using probabilities from $\textup{softmax}(\tau x)_{kl}$. 
      \State \textbf{Remove Edge:} for edge $(i, j)$ with maximal $\textup{Ric}(i, j)$ if $\textup{Ric}(i, j) > C^+$ and $T_{\textbf{v}_i \rightarrow \textbf{v}_j} < C^-$.
    \Until{Convergence or the maximum number of iterations is reached.}
  \end{algorithmic}
\end{algorithm}

In this work, \textit{we propose a new rule to calculate a vector for deciding a new edge and for deleting an edge by considering causal factors} in addition to the curvature called Causality-Informed Stochastic Discrete Ricci Flow (CSDRF). Our proposed rules for curvature graph rewiring are described in Algorithm~\ref{alg:sdrf}. The process requires the original causal graph's adjacency matrix $A$, temperature $\tau > 0$ to control sampling randomness, an optional Ricci curvature threshold $C^+$ (upper bound for edge removal), and an optional TE threshold $C^-$ (lower bound for edge removal). CSDRF consists of four steps: first, identifying the bottleneck edge by calculating Ricci curvature based on Equation (7). Second, computing improvement for candidate edges $(k, l)$, where $k \in B_1(i)$ and $l \in B_1(j)$, using $x_{kl} = (\textup{Ric}_{kl}(i, j) - \textup{Ric}(i, j)) \times T_{\textbf{v}_k \rightarrow \textbf{v}_l}$. Third, sampling a new edge based on $\textup{softmax}(\tau x)_{kl}$, where $\tau$ controls randomness, with higher $\tau$ making selection to be more deterministic (the best edge is always added). Finally, once a new edge is added, the algorithm removes the edge with maximal Ricci curvature $\textup{Ric}(i, j)$. The thresholds $C^+$ and $C^-$ ensure curvature remains balanced and strong causal relationships are preserved. At the end of the process, the refined causal graph is produced and used as input for the brain's spatial modeling.

\subsection{Brain's Spatial Modelling} We utilize a graph convolutional network (GCN)~\citep{welling2016semi_gcn} to capture spatial information from the generated brain structure. We use the refined causal graph developed from the brain's graph generation process as an input adjacency matrix $A$ into the GCN. The node features $X \in \mathbb{R}^{n \times n}$ represent the features of nodes, where $n$ corresponds to the number of nodes. These node features $X$ are derived from causal discovery based on TE calculations between ROIs. For example, a feature vector of node $i$ is $\textbf{x}_i \in \mathbb{R}^{n}$, which is derived from the TE calculation of ROI signal $\textbf{v}_i$ and any other ROI signals $\textbf{v}_j$ in a brain where $j=1,2, \ldots, n \text{ and } j \neq i$. Referring to~\cite{welling2016semi_gcn}, the GCN layer is defined as:

\begin{equation}
\label{eq:gcn}
H^{(\ell+1)} = \sigma (\hat{D}^{-\frac{1}{2}}\hat{A}\hat{D}^{-\frac{1}{2}}H^{(\ell)}\theta^{(\ell)}),
\end{equation}
where $H^{(\ell+1)}$ represents the new embedding of the overall node (ROI) representation, and $H^{(\ell)}$ denotes the embedding of the previous layer; $H^0 = X$. Additionally, $\hat{A}=A + I$ is a normalized adjacency matrix with self-loops added to the original graph. $\hat{D}$ stands for the diagonal degree matrix of $\hat{A}$, and $\theta^{(\ell)}$ represents the learnable parameters of layer $\ell$. The function $\sigma(\cdot)$ denotes the sigmoid function used for the linear model.

Furthermore, to improve the expressiveness of the node embedding after graph convolution, similar to the Vision GNN~\citep{han2022_visiongnn} concept, we implement the multi-head operation as a \textit{feature transform module}. We split the feature embedding of each node of $\textbf{h}^{(\ell+1)}_i$ from the overall node embedding matrix $H^{(\ell+1)}$ into $k$ heads, i.e., $head_1, head_2, ... head_k$. This strategy enables the projection of each node representation with different weights respectively. Then, all heads from the feature embedding can be updated in parallel, and the results are concatenated as the final values:

\begin{equation}
\begin{aligned}
\label{eq:multiHead}
\textbf{h}^{(\ell+1)}_i &= \mathbin\Vert_{j=1}^k head_j \theta_j \\
&= [head_1 \theta_1 \Vert head_2 \theta_2 \Vert \ldots \Vert head_k \theta_k],
\end{aligned}
\end{equation}

These multi-head operations enhance the feature diversity of the model. Next, to prevent the vanishing gradient problem in each layer of the brain's spatial modeling, we introduce a skip connection so that the newly updated node is written as: $\textbf{h}^{(\ell+1)}_i = \sigma(\textbf{h}_i^{(\ell+1)}) + \textbf{h}_i^{(\ell)}$. Here, $\textbf{h}^{(\ell+1)}_i$ is the newly updated node $i$ representation after the multi-head operation, $\textbf{h}^{(\ell)}$ is the previous node $i$ representation before the GCN operation, and $\sigma(\cdot)$ is an activation function such as ReLU.

In this study, the representation of node embedding $H$ after undergoing $\ell$ layers of convolutional operations is denoted as $Z \in \mathbb{R}^{n \times f}$, where $f$ represents the dimension of the node embedding. The new overall nodes (ROIs) embedding $Z$ will be modeled to summarize the general brain network representation by aggregating nodes into a concise, high-level picture.

\subsection{Pooling and Classifier}

This module summarizes the obtained node embedding $Z$ from the previous step of the brain's spatial modeling into the graph-level embedding $\textbf{z}^{\textup{pooled}}$. The pooling strategy is to concatenate node features of all nodes (ROIs) in the graph. For each brain structure, CONCAT pooling is performed as follows:

\begin{equation}
  \label{eq:concat}
   \textbf{z}^{\textup{pooled}} = \mathbin\Vert_{j=1}^n \textbf{z}_j = [\textbf{z}_1 \Vert \textbf{z}_2 \Vert \ldots \Vert \textbf{z}_n],
\end{equation}
where $\textbf{z}^{\textup{pooled}} \in \mathbb{R}^{m}$ represents the graph-level embedding with size $m=f\times n$ and $ \textbf{z}_j \in \mathbb{R}^{f}$ is each node embedding in the graph $G$. 

Once we obtain the graph representation $\textbf{z}^{\textup{pooled}}$, we then apply double fully connected layers as a \textit{feature transform module} to implement binary classification for brain disease or normal samples, that is, $y = \textup{FC}(\textup{FC}(\textbf{z}^{\textup{pooled}}))$. Here, $\textup{FC}(\cdot)$ represents a fully connected network, and the output $y$ of this transformation is mapped to $1$, indicating a specific brain disease, or $0$ for normal samples.

\section{Experiments}
In this section, we will explain the experiments we conducted to address the following research objectives:

\begin{itemize}
    \item \textbf{RQ1}: Can CGB perform better in brain disease detection/classification tasks compared to the benchmark methods?
    \item \textbf{RQ2}: Does each module in CGB substantially influence the improvement of brain disease classification performance?
    \item \textbf{RQ3}: What is the sensitivity of the important control parameter in the causal graph generation given different values?
    \item \textbf{RQ4}: How to interpret the causal graph structure as a brain network in the CGB framework?
    
\end{itemize}

\begin{table}[!ht]
\centering
\caption{Statistics of datasets.}
\label{tb:dataset}
\begin{tabular}{cccc}
\hline
\textbf{Datasets} & \textbf{Nodes/ROIs} & \textbf{Diseased Samples} & \textbf{Healthy Samples} \\ \hline
COBRE & 96 & 72 & 75 \\
ACPI & 200 & 62 & 64 \\
ABIDE & 111 & 402 & 464 \\ 
ADNI & 90 & 54 & 211 \\\hline
\end{tabular}%
\end{table}

\subsection{Datasets}
We conducted experiments using four open-source fMRI brain datasets:
\begin{itemize}
    \item COBRE\footnote{http://cobre.mrn.org/}: The COBRE dataset was developed by the Center for Biomedical Research Excellence. It includes fMRI data from 147 patients, with 72 diagnosed with schizophrenia and 75 as healthy controls. This dataset was preprocessed using the Harvard-Oxford parcellation technique~\citep{makris2006decreased_harvard_Parcellation}, which consists of 96 regions of interest (ROIs), each represented by a temporal signal spanning 150-time steps.
    \item ACPI\footnote{http://fcon\_1000.projects.nitrc.org/indi/ACPI/html/}: The Addiction Connectome Preprocessed Initiative (ACPI) dataset is supported by the National Institute on Drug Abuse. It contains fMRI data from 128 patients, with 62 diagnosed with Attention-deficit/hyperactivity disorder (ADHD) with marijuana consumption and 64 as healthy controls. This dataset was preprocessed using the Craddock 200 (CC200) parcellation technique~\citep{craddock2012whole_cradock_Parcellation} based on the original source of the dataset, consisting of 200 ROIs, with each ROI comprising a temporal signal of 700 time steps.
    \item ABIDE~\citep{di2014autism_abide} The Autism Brain Imaging Data Exchange (ABIDE) dataset was collected from the preprocessed connectome project (PCP)~\citep{craddock2013neuro_PCP}. It includes fMRI data from 868 patients, with 402 diagnosed with autism and 464 as healthy controls. This dataset was preprocessed using Harvard-Oxford parcellation technique~\citep{makris2006decreased_harvard_Parcellation} for the ABIDE dataset, which consists of 11 ROI nodes, and each ROI consists of 192 time steps of temporal signals.
    \item ADNI\footnote{https://adni.loni.usc.edu/}: For ADNI (Alzheimer's Disease Neuroimaging Initiative), we use a total of 265 samples from patients aged 57–93 years, consisting of 211 normal control (NC) patients and 54 patients with Alzheimer’s Disease (AD). ADNI uses Automated Anatomical Labeling (AAL) parcellation technique, which contains 90 ROIs.
\end{itemize}

It is important to note that the use of different parcellation techniques was determined by the unique parameters and sampling equipment associated with each dataset. The specific atlas developed for each dataset indicates that it is better suited for the recovery of fMRI signals, reflecting the unique characteristics of the data~\citep{li2024individual_different_atlas}. A summary of the dataset statistics is provided in Table~\ref{tb:dataset}. Nodes/ROIs represent the number of nodes in brain networks. Diseased samples indicate the number of data points associated with specific disease labels, while healthy samples refer to data points representing normal observations.

\subsection{Baseline Methods}
We compare the performance of our proposed framework with current brain disease classification frameworks such as BrainNetCNN~\citep{kawahara2017brainnetcnn}, BrainGNN~\citep{li2021braingnn}, BrainGB~\citep{cui2022braingb}, GDC-GCN~\citep{elgazzar2022benchmarking}, DGM~\citep{kazi2022differentiable_dgm}, FBNetGNN~\citep{kan2022fbnetgen}, BrainNETTF~\citep{kan2022brain_brain_transformer} and Bargrain~\citep{febrinanto2023balanced}. We used their original code implementations to carry out the experiment.

\subsection{Evaluation Metrics}
In this work, all datasets are used for prediction tasks based on binary classification problems. We evaluate all models, including ours, with common classification evaluation metrics such as F1 score and Area Under the ROC Curve (AUC). Since the models are for medical applications, we also add two other metrics for diagnostic tests, which are sensitivity and specificity. Sensitivity defines the true positive rate, and specificity refers to the true negative rate.

\subsection{Experimental Settings}

This section explains the implementation settings for conducting the experiments. For all datasets, we divided each dataset into 80\% for training sets and 20\% for testing sets. We also split the training sets into 85\% for actual training sets and 15\% for validation sets to aid in model selection.

We conducted the experiments using an AMD Ryzen 7 5800H @ 3.20 GHz processor and an NVIDIA GeForce RTX 3050 Ti GPU. For graph embedding and feature transformation in brain spatial modeling, we applied a pyramid architecture~\citep{han2022_visiongnn} with progressively halved embedding sizes per layer. The optimal hidden embedding size of brain's spatial modelling, \textit{determined during model selection based on validation sets}, was set as a hyperparameter: 256 for COBRE and 128 for ACPI, ABIDE, and ADNI datasets. In the brain's spatial modeling module, we used 4 layers for COBRE, ABIDE, and ADNI, and 2 for ACPI. For multi-head feature transformation, we applied 2 heads for COBRE and 4 for ACPI, ABIDE, and ADNI.

The pooling and classifier modules have two types of hidden dimensions: one for the first pooling feature transformation layer and another for the second. For the COBRE dataset, the dimensions are 256 and 128 for the first and second layers, respectively. The ACPI dataset uses 512 for the first layer and 64 for the second, while the ABIDE dataset uses 128 for both layers. Similarly, the ADNI dataset has a hidden dimension of 256 for both layers.

To maintain the sparsity of the causal graph, the value of the $c$ parameter should be selected based on the dataset's characteristics. A grid search on validation subsets should be performed to identify the optimal $c$ based on the classification performance. Start with $c \approx 0$ to eliminate weak or negative causal edges and incrementally adjust upward to focus the graph on stronger causal relationships. When no target labels are available, domain knowledge may be required, though this is beyond the scope of our framework due to its supervised learning settings. In this work, we set a control variable $c$ to 0.1 for the COBRE, ABIDE and ADNI datasets and 0.05 for the ACPI dataset. 

We applied a 0.5 dropout rate during brain spatial modeling to prevent overfitting. Training used the Adam optimizer with a 0.0001 learning rate, a batch size of 16, and up to 200 epochs. Early stopping monitored validation performance, stopping training if no improvement was observed.

\begin{table*}[!ht]
  \centering
  \caption{Experimental results over four datasets. The \textbf{bold} text indicates the best results.}
  \label{tb:results}
  \resizebox{\textwidth}{!}{
\begin{tabular}{clllllllll}
\hline
\multicolumn{1}{c|}{\multirow{2}{*}{\textbf{Graph Types}}} & \multicolumn{1}{c|}{\multirow{2}{*}{\textbf{Models}}} & \multicolumn{4}{c|}{\textbf{COBRE}} & \multicolumn{4}{c}{\textbf{ACPI}} \\ 
\multicolumn{1}{c|}{} & \multicolumn{1}{c|}{} & \multicolumn{1}{c}{\textbf{F1}} & \multicolumn{1}{c}{\textbf{Sens.}} & \multicolumn{1}{c}{\textbf{Spec.}} & \multicolumn{1}{c|}{\textbf{AUC}} & \multicolumn{1}{c}{\textbf{F1}} & \multicolumn{1}{c}{\textbf{Sens.}} & \multicolumn{1}{c}{\textbf{Spec.}} & \multicolumn{1}{c}{\textbf{AUC}} \\ \hline
\multicolumn{1}{c|}{Non-Graph} & \multicolumn{1}{l|}{BrainNetCNN} & 0.6923 & 0.6429 & \textbf{0.8000} & \multicolumn{1}{l|}{0.7214} & 0.5833 & 0.5833 & 0.6154 & 0.5994 \\ \hline
\multicolumn{1}{c|}{\multirow{3}{*}{\begin{tabular}[c]{@{}c@{}}Fixed \\ Correlation \\ Graph\end{tabular}}} & \multicolumn{1}{l|}{BrainGNN} & 0.3846 & 0.3571 & 0.5333 & \multicolumn{1}{l|}{0.4452} & 0.6310 & \textbf{0.8333} & 0.2538 & 0.5436 \\
\multicolumn{1}{c|}{} & \multicolumn{1}{l|}{BrainGB} & 0.6000 & 0.6429 & 0.5333 & \multicolumn{1}{l|}{0.5881} & 0.4545 & 0.4167 & 0.6154 & 0.5160 \\
\multicolumn{1}{c|}{} & \multicolumn{1}{l|}{GDC-GCN} & 0.5185 & 0.5000 & 0.6000 & \multicolumn{1}{l|}{0.5500} & 0.6250 & \textbf{0.8333} & 0.2308 & 0.5321 \\ \hline
\multicolumn{1}{c|}{\multirow{3}{*}{\begin{tabular}[c]{@{}c@{}}Learnable\\ Graph\end{tabular}}} & \multicolumn{1}{l|}{DGM} & 0.6400 & 0.5714 & \textbf{0.8000} & \multicolumn{1}{l|}{0.6857} & 0.7200 & 0.7500 & 0.6923 & 0.7212 \\
\multicolumn{1}{c|}{} & \multicolumn{1}{l|}{FBNetGNN} & 0.5185 & 0.5000 & 0.6000 & \multicolumn{1}{l|}{0.5500} & 0.5600 & 0.5833 & 0.5385 & 0.5609 \\
\multicolumn{1}{c|}{} & \multicolumn{1}{l|}{BrainNETTF} & 0.5217 & 0.4286 & \textbf{0.8000} & \multicolumn{1}{l|}{0.6143} & 0.6400 & 0.6667 & 0.6154 & 0.6410 \\ \hline
\multicolumn{1}{c|}{Mixed Graph} & \multicolumn{1}{l|}{Bargrain} & 0.7407 & 0.7143 & \textbf{0.8000} & \multicolumn{1}{l|}{0.7571} & 0.7407 & \textbf{0.8333} & 0.6154 & 0.7244 \\ \hline
\multicolumn{1}{c|}{Ours} & \multicolumn{1}{l|}{\textbf{CGB}} & \textbf{0.7586} & \textbf{0.7857} & 0.7333 & \multicolumn{1}{l|}{\textbf{0.7595}} & \textbf{0.7619} & 0.6667 & \textbf{0.9231} & \textbf{0.7949} \\ \hline
\multicolumn{1}{l}{} &  &  &  &  &  &  &  &  &  \\ \hline
\multicolumn{1}{c|}{\multirow{2}{*}{\textbf{Graph Types}}} & \multicolumn{1}{c|}{\multirow{2}{*}{\textbf{Models}}} & \multicolumn{4}{c|}{\textbf{ABIDE}} & \multicolumn{4}{c}{\textbf{ADNI}} \\ 
\multicolumn{1}{c|}{} & \multicolumn{1}{c|}{} & \multicolumn{1}{c}{\textbf{F1}} & \multicolumn{1}{c}{\textbf{Sens.}} & \multicolumn{1}{c}{\textbf{Spec.}} & \multicolumn{1}{c|}{\textbf{AUC}} & \multicolumn{1}{c}{\textbf{F1}} & \multicolumn{1}{c}{\textbf{Sens.}} & \multicolumn{1}{c}{\textbf{Spec.}} & \multicolumn{1}{c}{\textbf{AUC}} \\ \hline
\multicolumn{1}{c|}{Non-Graph} & \multicolumn{1}{l|}{BrainNetCNN} & 0.7158 & 0.7234 & \textbf{0.6543} & \multicolumn{1}{l|}{0.6889} & 0.4706 & 0.3636 & \textbf{0.9524} & 0.8333 \\ \hline
\multicolumn{1}{c|}{\multirow{3}{*}{\begin{tabular}[c]{@{}c@{}}Fixed \\ Correlation \\ Graph\end{tabular}}} & \multicolumn{1}{l|}{BrainGNN} & 0.6599 & 0.7213 & 0.4605 & \multicolumn{1}{l|}{0.5909} & 0.2143 & 0.2727 & 0.6667 & 0.4589 \\
\multicolumn{1}{c|}{} & \multicolumn{1}{l|}{BrainGB} & 0.6288 & 0.6266 & 0.5840 & \multicolumn{1}{l|}{0.6053} & 0.5143 & 0.8182 & 0.6429 & 0.7814 \\
\multicolumn{1}{c|}{} & \multicolumn{1}{l|}{GDC-GCN} & 0.6377 & 0.7021 & 0.4198 & \multicolumn{1}{l|}{0.5609} & 0.3158 & 0.2727 & 0.8810 & 0.5693 \\ \hline
\multicolumn{1}{c|}{\multirow{3}{*}{\begin{tabular}[c]{@{}c@{}}Learnable\\ Graph\end{tabular}}} & \multicolumn{1}{l|}{DGM} & 0.5631 & 0.6170 & 0.3333 & \multicolumn{1}{l|}{0.4752} & 0.3529 & 0.2727 & 0.9286 & 0.7662 \\
\multicolumn{1}{c|}{} & \multicolumn{1}{l|}{FBNetGNN} & 0.6486 & 0.6383 & 0.6173 & \multicolumn{1}{l|}{0.6278} &  0.4444 & 0.3636 & 0.9286 & 0.8139 \\
\multicolumn{1}{c|}{} & \multicolumn{1}{l|}{BrainNETTF} & 0.6984 & 0.7021 & 0.6420 & \multicolumn{1}{l|}{0.6721} & 0.4878 & \textbf{0.9091} & 0.5238 &  0.8247 \\ \hline
\multicolumn{1}{c|}{Mixed Graph} & \multicolumn{1}{l|}{Bargrain} & 0.7172 & 0.7553 & 0.5926 & \multicolumn{1}{l|}{0.6740} & 0.5217 & 0.5455 & 0.8571 & 0.7727 \\ \hline
\multicolumn{1}{c|}{Ours} & \multicolumn{1}{l|}{\textbf{CGB}} & \textbf{0.7449} & \textbf{0.7766} & 0.6420 & \multicolumn{1}{l|}{\textbf{0.7093}} & \textbf{0.5882} & \textbf{0.9091} & 0.6905 & \textbf{0.8593} \\ \hline
\end{tabular}
}
\end{table*}

\begin{table*}[ht!]
\centering
\caption{Averages of F1 scores (Avg. F1) of all methods. The \textbf{bold} text indicates the best results.}
\label{tb:f1_results}
\resizebox{0.45\textwidth}{!}{
\begin{tabular}{c|l|l}
\hline
\textbf{Graph Types} & \multicolumn{1}{c|}{\textbf{Models}} & \multicolumn{1}{c}{\textbf{Avg. F1}} \\ \hline
Non-Graph & BrainNetCNN & 0.6155 \\ \hline
\multirow{3}{*}{\begin{tabular}[c]{@{}c@{}}Fixed \\ Correlation \\ Graph\end{tabular}} & BrainGNN & 0.4725
 \\
 & BrainGB & 0.5494 \\
 & GDC-GCN & 0.5243 \\ \hline
\multirow{3}{*}{\begin{tabular}[c]{@{}c@{}}Learnable\\ Graph\end{tabular}} & DGM & 0.5690 \\
 & FBNetGNN & 0.5429 \\
 & BrainNETTF & 0.5870 \\ \hline
Mixed Graph & Bargrain & 0.6801 \\ \hline
Ours & \textbf{CGB} & \textbf{0.7134} \\ \hline
\end{tabular}}
\end{table*}

\subsection{RQ1: Experimental Results}
The experimental results, comparing our CGB framework with other recent brain disease classification frameworks, are shown in Table~\ref{tb:results}. The CGB framework outperforms other benchmark models, achieving an average F1 score of 0.7134 as summarized in Table~\ref{tb:f1_results}. Unlike other frameworks that use correlation graphs, such as GDC-GCN~\citep{elgazzar2022benchmarking}, BrainGB~\citep{cui2022braingb}, BrainGNN~\citep{li2021braingnn}, and Bargrain~\citep{febrinanto2023balanced}, CGB employs refined causal graphs to represent brain networks. Relationships over the brains' ROIs in the CGB framework help improve classification performance since it brings an understanding of cause-and-effect relationships in brain networks, leading to more robust predictions.

Similarly, learnable graph techniques such as DGM~\citep{kazi2022differentiable_dgm}, FBNetGNN~\citep{kan2022fbnetgen}, and BrainNETTF~\citep{kan2022brain_brain_transformer}, which focus on identifying optimal graph structures, could not achieve better classification performance. This is due to some limitations of learnable graph techniques that rely on training samples for graph construction, which can lead to easy overfitting and poor generalization on test data where some data were unseen before. In contrast, the causality discovery technique in CGB can incorporate domain knowledge structure based on actual biological insights, thus improving the models' generalization. Moreover, the graph refinement process based on geometric curvature also helps to enhance the expressiveness of brain graph structures to be better modeled by GNNs. Therefore, the findings suggest that using a causal graph structure as a brain network can enhance classification quality in distinguishing healthy and diseased brains.

\begin{figure}[!htb]
\centering
\includegraphics[width=0.75\textwidth]{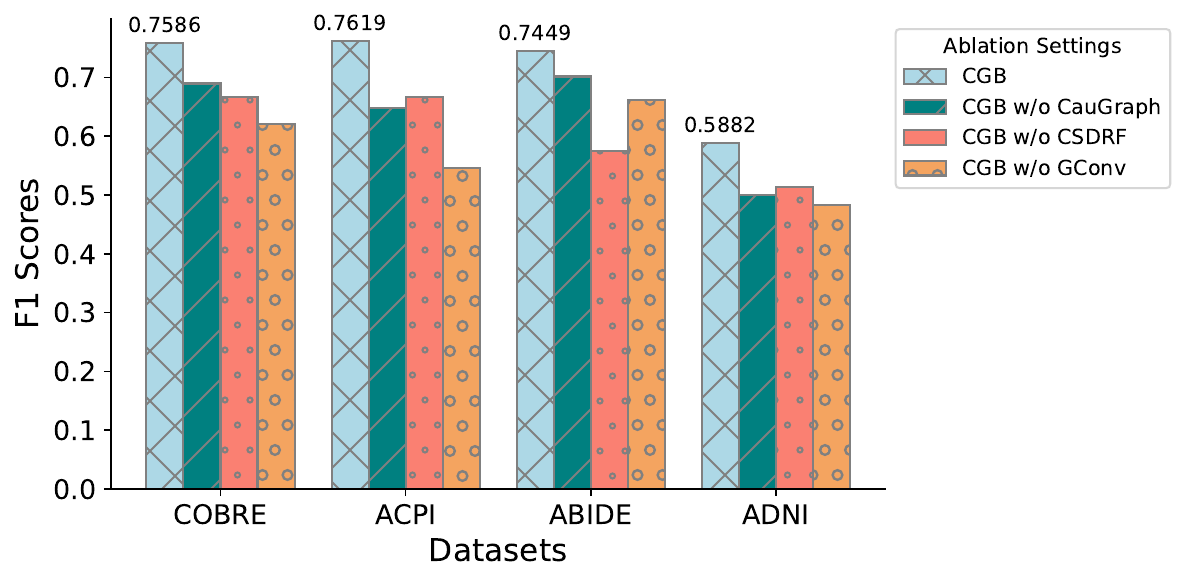}
\caption{F1 scores under different ablation settings.}
\label{fig:ablations}
\end{figure}

\subsection{RQ2: Ablation Studies}
In this section, we evaluate the impact of each component in our model through ablation studies, where we systematically remove specific elements and observe the results. We consider three settings: First, we exclude both the causal graph and the graph refinement strategy (\textit{CGB w/o CauGraph}), replacing them with a thresholded correlation matrix graph. Second, we remove our graph rewiring techniques based on geometric curvature, specifically the Causality-Informed Stochastic Discrete Ricci Flow (CSDRF) (\textit{CGB w/o CSDRF}), and instead use thresholded causal graphs. Finally, in the third setting, we eliminate the graph embedding process, running the framework without graph convolution (\textit{CGB w/o GConv}).

The result of ablation studies in Figure~\ref{fig:ablations} shows the effectiveness of the complete framework of CGB. Classification performance decreases when the causal graph as a brain network is not used. It suggests that the causality factor between the brain's ROI has a better representation to distinguish normal and diseased brains. Using geometry curvature can improve the accuracy performance, making the graph structure more expressive and reducing potential noise from causality discovery calculation. Moreover, using graph convolution to model graph data reveals the high-level characteristics of the refined causal graph into a meaningful representation, which also helps the classification process.

\begin{figure}[!ht]
  \centering
  \includegraphics[width=0.65\textwidth]{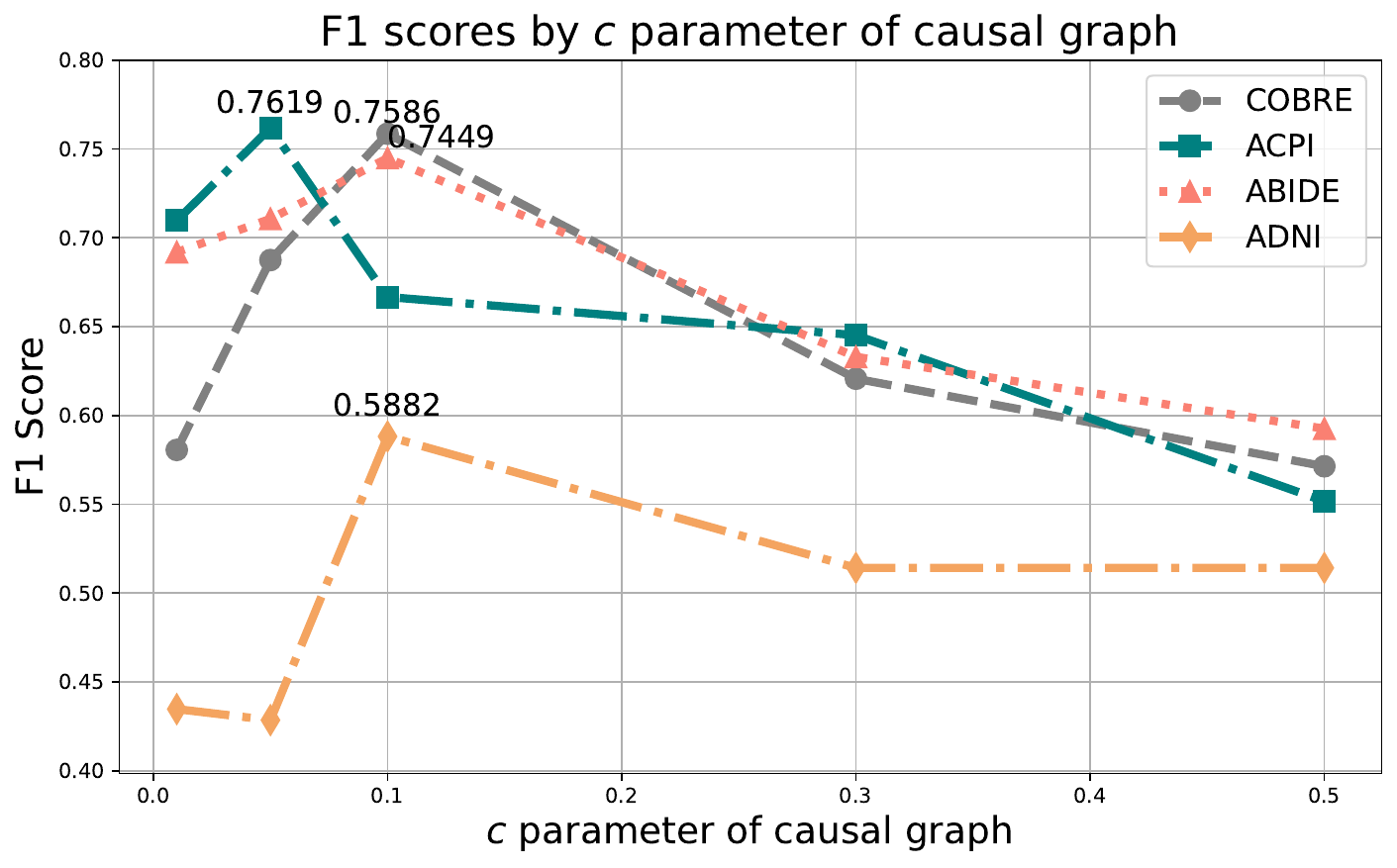}
  \caption{Comparison of F1 scores across various values of 
  $c$ for graph structure utilization.}
  \label{fig:sensitivity}
  \end{figure}
  
\subsection{RQ3: Causal Graphs' Control Constant Sensitivity Analysis}
\label{sec:sens}
In this section, we study the sensitivity of the important variable, control constant $c$, in Equation~\ref{eq:adj}. The control constant $c$ parameter aims to control the sparsity of the generated causal graph before the refinement process. The lower value of $c$ indicates that the graph will consider most of the strong causal and weak causal relationships. The higher value of $c$ will filter out the weak causal relationships to focus on the stronger ones.

The result of parameter sensitivity of this parameter is shown in Fig.~\ref{fig:sensitivity}. We examine different values of $c$, such as 0.01, 0.05, 0.1, 0.2, 0.3, 0.4, and 0.5 for all datasets. The result shows that the suitable value of the $c$ parameter for the dataset ACPI, ABIDE and ADNI is 0.1, and for the COBRE dataset is 0.05. The result indicates that the preferable choice for the $c$ parameter should be lower around 0.1. Increasing the value of $c$ will remove some edges that bring causal information. Avoiding a value close to 0 could also help improve the performance by eliminating very few weak causal relationships.

\subsection{RQ4: Causal Graph Interpretation}
This section presents an interpretation of generated causal graphs in brain networks. It answers whether the causal graph generation in CGB can discover relevant and intrinsic relationships between the brain's ROIs.

\begin{figure}[!ht]
  \centering
  \includegraphics[width=0.5\textwidth]{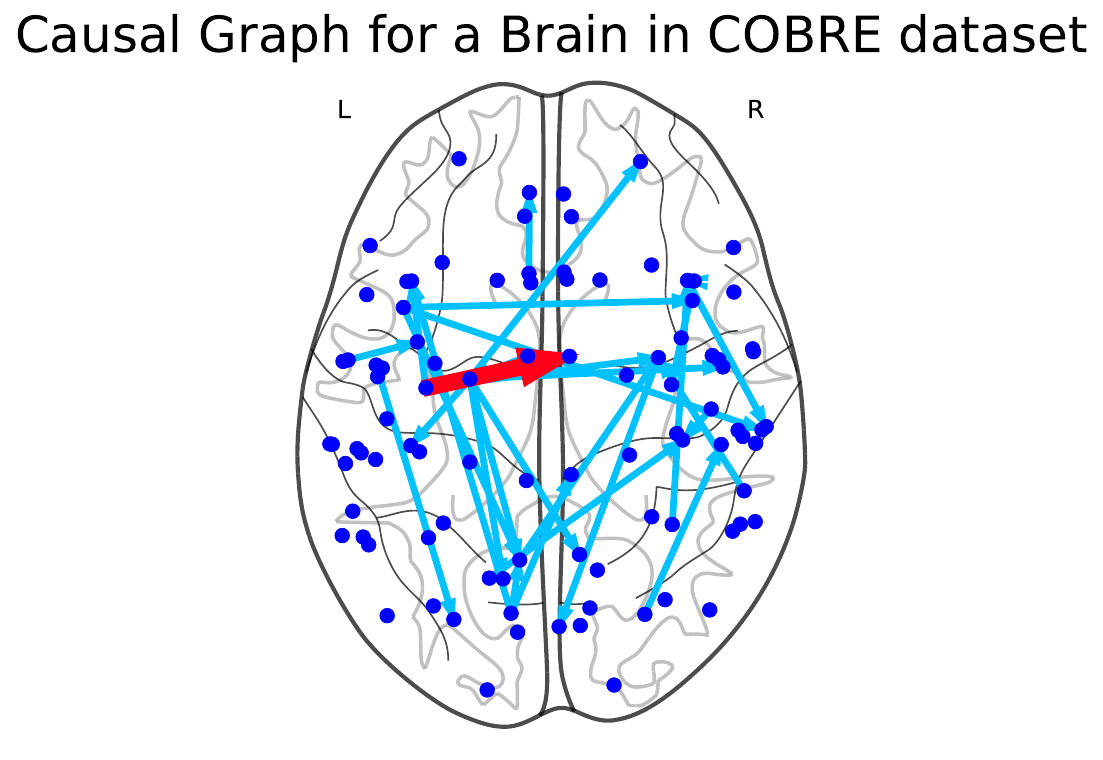}
  \caption{A causal graph for a brain from the COBRE dataset. The red arrow represents the strongest causal factor in the graph, indicating the direction from Node 51, termed \textit{Left Juxtapositional Lobule Cortex}, to Node 12, termed \textit{Right Inferior Frontal Gyrus}, denoted as $T_{\textbf{v}_{51}\rightarrow \textbf{v}_{12}}$.
  }
  \label{fig:brain}
\end{figure}

\begin{figure}[!ht]
  \centering
  \includegraphics[width=\textwidth]{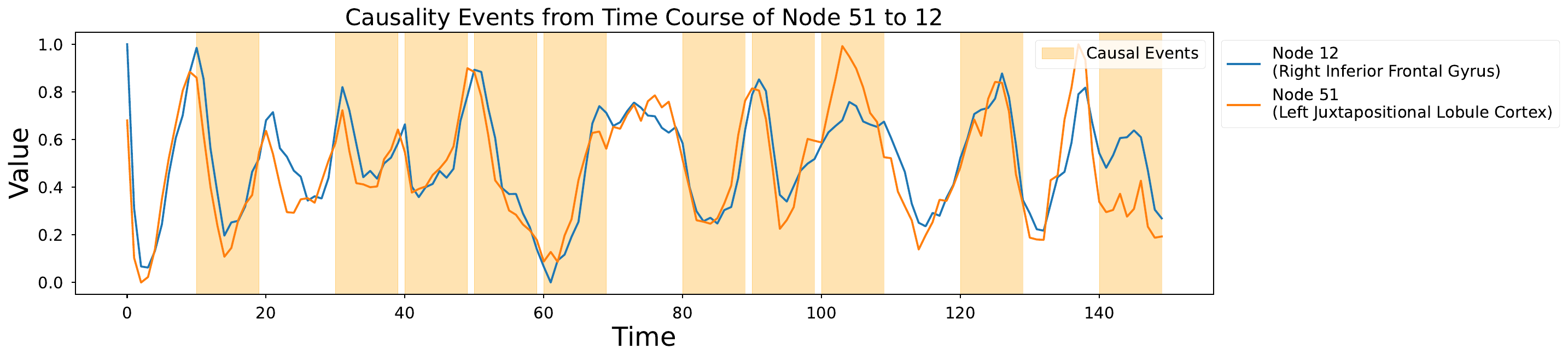}
  \caption{Causality events of fMRI time courses between Node 51, termed \textit{Left Juxtapositional Lobule Cortex}, and Node 12, termed \textit{Right Inferior Frontal Gyrus}, ($T_{\textbf{v}_{51}\rightarrow \textbf{v}_{12}}$), in a brain from the COBRE dataset. These two nodes have the highest causality relationship, as shown in Fig.~\ref{fig:brain}.}
  \label{fig:causal}
\end{figure}

\begin{figure}[!ht]
  \centering
  \includegraphics[width=\textwidth]{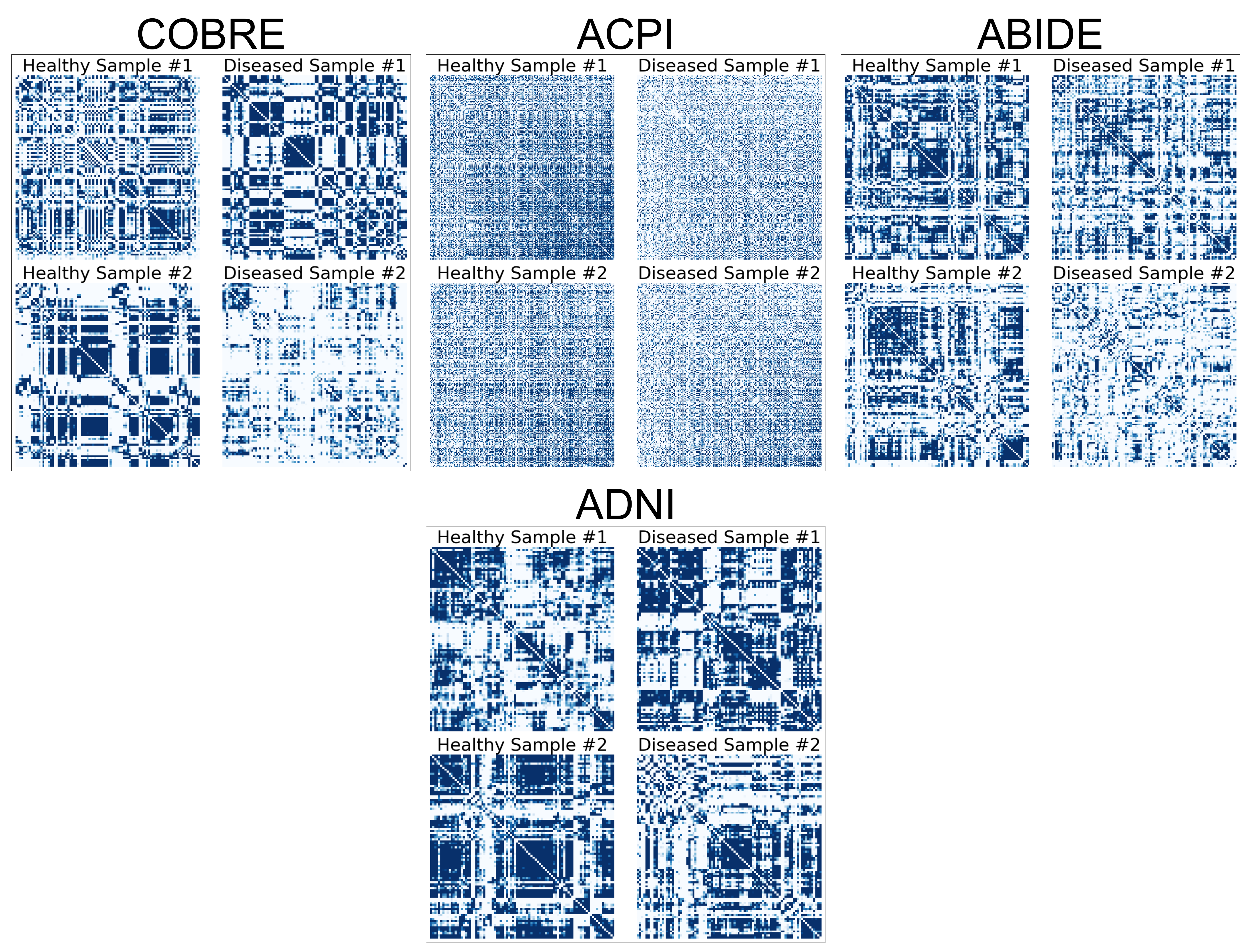}
  \caption{Heatmap of Causal Relationships in Healthy and Diseased Samples Across Datasets.}
  \label{fig:causal_graphs}
\end{figure}

Starting with Fig.~\ref{fig:brain}, we visualize the generated causal graph in a brain. To help this visualization, we choose one brain sample from the COBRE dataset. Based on Fig.~\ref{fig:brain}, the blue dots represent the nodes or brain's ROIs. There are 96 nodes/ROIs in each sample of the COBRE dataset with different terms and positions. In this part, we only choose the top 1\% edges from the generated causal graph adjacency matrix for visualization purposes. The blue light arrows represent the top 1\% edges, and the red arrow represents the strongest causal relationship in that brain sample. From the causal discovery calculation, it concludes that the highest causal relationship flow from Node 51, called \textit{Left Juxtapositional Lobule Cortex}, to Node 12, called \textit{Right Inferior Frontal Gyrus} with transfer entropy score $T_{\textbf{v}_{51}\rightarrow \textbf{v}_{12}}$, 0.6191.

Next, after knowing the highest causality score within the graph, we visualize the causal events in Fig.~\ref{fig:causal}. We do it by calculating the transfer entropy of those two fMRI time courses of nodes 51 and 12 on every 10-time point. Furthermore, we highlight the top 10 causal events from each group of 10-time points to show the interpretation of high causality. From that visualization, it is noticeable that when two signals have a high causality factor, it has a strong intrinsic relationship, such as when some values decreased in node 51 and node 12 also decreased in the following timeframe. Similarly, when the value of node 51 increases, then right after that, the value of node 12 also increases. From overall visualization of causal graph interpretation and the result of the ablation study in Fig.~\ref{fig:ablations} demonstrate how the CGB framework improves the classification quality while introducing intrinsic relationships to represent the brain networks. 

\textbf{Heatmap of causal relationships}. We also visualize the heatmap of causal relationships in healthy and diseased samples across all datasets, as shown in Figure~\ref{fig:causal_graphs}. The causal discovery mechanism uncovers non-trivial relationships, with stronger causality represented by darker spots. These meaningful relationships are useful for developing the graphs as input for performing graph-level classification. However, this visualization alone is insufficient for distinguishing between healthy and diseased samples. While differences between these groups may be present, they are often subtle and not consistently clear. This underscores the advantage of deep learning algorithms, particularly GNNs, which employ a \textit{black-box} mechanism~\citep{zhou2022graph_black} capable of uncovering complex patterns by analyzing not only the graph structures but also the features within each node.

\section{Discussion and Future Directions}
While the CGB demonstrated improved performance in brain disease classification through its refined causal graph and brain spatial modeling, some limitations remain. For instance, the ADNI dataset results revealed lower performance, largely due to the imbalance between healthy and diseased samples. This issue also affects baseline models, which similarly struggle under such conditions. Addressing the challenge of unbalanced samples could be a valuable direction for future research. Additionally, this work relies on supervised learning, requiring labeled data, which may be limited in certain contexts. Exploring semi-supervised or unsupervised learning techniques could help overcome the challenges posed by low-label availability in future studies.

Another promising direction for future work is to investigate dynamic causal graphs. Causal relationships might be dynamic. By segmenting time series signals from different ROIs into distinct snapshots~\citep{tieu2025temporal_dynamic} and computing causal discovery measures for each, more diverse causal connections could be captured. However, this approach would require iterative graph rewiring and a careful trade-off between computational complexity and classification performance. These challenges must be addressed to enable effective and efficient dynamic refined causal graphs, making this a promising direction for future work in modeling complex brain dynamics.

\section{Conclusion}
In this work, we propose CGB (Causal Graphs for Brains) to perform brain disease classification. Our CGB framework employs the brain's graph generation based on causal discovery based on the TE technique and graph refinement strategy based on geometric curvature. The causality discovery aims to build a graph adjacency matrix, and graph refinement seeks to improve the expressiveness of the graph structures, reducing potential noise from causal discovery calculation and preventing information bottlenecks while modeling it using GNNs. Our framework demonstrates an excellent performance through extensive experiments compared to the recent brain disease classification techniques. Apart from several future directions discussed in the experiments section, our future work will also explore the causality factor among different modalities, not only from fMRI signals but also from other signals, such as diffusion MRI (dMRI) or electroencephalography (EEG), to improve classification performance.

\section*{Data availability statement}
The authors declare that the datasets supporting experiments and findings of this study are available in public repositories. The COBRE dataset developed by the Center for Biomedical Research Excellence is available at https://fcon\_1000.projects.nitrc.org/indi/retro/cobre.html. The Addiction Connectome Preprocessed Initiative (ACPI) dataset supported by the National Institute on Drug Abuse is available at https://fcon\_1000.projects.nitrc.org/indi/ACPI/html/acpi\_mta\_1.html. The Autism Brain Imaging Data Exchange (ABIDE) is available at https://doi.org/10.1038/mp.2013.78. The Alzheimer’s Disease Neuroimaging Initiative (ADNI) dataset is available at https://adni.loni.usc.edu/about/.








\bibliographystyle{sn-basic.bst} 
\bibliography{bib.bib}

\hspace{1cm}

\textbf{Falih Gozi Febrinanto} received his bachelor of computer science degree from the University of Brawijaya, Malang, Indonesia, in 2018 and his master of technology degree from Federation University Australia, Ballarat, Australia, in 2021. He is currently pursuing a Ph.D. degree in information technology at Federation University Australia, Ballarat, Australia. His research interests include graph learning, artificial intelligence, and anomaly detection. \\

\textbf{Adonia Simango} earned his Bachelor of Mathematics degree from Midlands State University in Zimbabwe in 2006, followed by a Master’s degree in Industrial Mathematics from the National University of Science and Technology, also in Zimbabwe, in 2008. Since then, he has worked in South Africa in various industry roles including Statistician, Analyst, and for the past six years, as a Data Scientist. His career spans several sectors such as public healthcare, retail consumer goods, consulting, and mining. His research interests include graph learning, data science, artificial intelligence, anomaly detection, ethical AI, and blockchain technology.\\

\textbf{Chengpei Xu} obtained his bachelor's degree in system engineering from the National University of Defence Technology, China, in 2014, his master's degree in computer science from the University of New South Wales, Australia, in 2018, and his Ph.D. in information technology from the University of Technology Sydney, Australia, in 2022. Presently, he serves as a Postdoctoral Research Fellow with MIoT \& IPIN Lab, School of Minerals and Energy Resources Engineering, University of New South Wales, NSW, 2052, Australia. His research areas encompass low-level computer vision, scene text detection and recognition,  and multimedia content understanding and generation.\\

\textbf{Jingjing Zhou} received the M.S. degree from Shandong University in 2004 and the Ph.D. degree from University of Science and Technology of Beijing in 2009. She is currently Associate Professor in School of Information and Electronic Engineering, Zhejiang Gongshang University, China. Her research interests include artificial intelligence, graph learning, and computer networks. \\

\textbf{Jiangang Ma} is a senior lecturer of information technology with Federation University Australia. He earned his PhD degree in computer science from Victoria University, Australia. His research interests include data science, algorithms design, artificial intelligence, image/video processing, internet of things, and health informatics. He has published over 50 research papers on international journals and conferences, including the papers published on top international journals and conferences, such as IEEE International Conference on Data Engineering (ICDE) and IEEE International Conference on Data Mining (ICDM), etc.\\

\textbf{Sonika Tyagi} is an Associate Professor of Digital Health and Bioinformatics at the School of Computational Technologies, RMIT University Australia. She is also an affiliate Machine Learning lead scientist at Central Clinical School Monash University Australia. Sonika's expertise is in developing new machine learning tools and pipelines and applying these methods to solve biological and clinical research questions. Sonika's current research focuses on integrative approaches to digital health and genomics.\\

\textbf{Feng Xia} received the BSc and PhD degrees from Zhejiang University, Hangzhou, China. He is a Professor in School of Computing Technologies, RMIT University, Australia. Dr. Xia has published over 300 scientific papers in journals and conferences (such as IEEE TAI, TKDE, TNNLS, TC, TMC, TBD, TCSS, TNSE, TETCI, TETC, THMS, TVT, TITS, ACM TKDD, TIST, TWEB, TOMM; IJCAI, AAAI, NeurIPS, ICLR, KDD, WWW, MM, SIGIR, EMNLP, and INFOCOM). His research interests include artificial intelligence, graph learning, brain, robotics, and cyber-physical systems. He is a Senior Member of IEEE and ACM, and an ACM Distinguished Speaker.

\end{document}